\documentclass[11pt]{article}
\usepackage{textcomp}

\usepackage[final]{acl}

\usepackage{times}
\usepackage{latexsym}
\usepackage{amsmath}
\usepackage{amssymb}
 \usepackage{booktabs} 
\usepackage{algorithm}
\usepackage{algorithmic}
\usepackage{float}  
\usepackage{amsmath} 
\usepackage{upgreek}
\usepackage[T1]{fontenc}

\usepackage[utf8]{inputenc}

\usepackage{microtype}

\usepackage{inconsolata}

\usepackage{graphicx}

\title{VideoChain: A Transformer-Based Framework for Multi-hop Video Question Generation}
\author{Arpan Phukan, Anupam Pandey, Deepjyoti Bodo \and Asif Ekbal \\
        Department of Computer Science and Engineering, IIT Patna \\ 
\texttt{arpan\_2121cs33@iitp.ac.in/iitpainlpmlresourcerequest@gmail.com,}\\
\texttt{anupam\_2311cs31@iitp.ac.in, deepjyoti\_2311ai65@iitp.ac.in, asif@iitp.ac.in}}

\begin{document}
\maketitle
\begin{abstract}
 Multi-hop Question Generation (QG) effectively evaluates reasoning but remains confined to text; Video Question Generation (VideoQG) is limited to zero-hop questions over single segments. 
 To address this, we introduce VideoChain, a novel Multi-hop Video Question Generation (MVQG) framework designed to generate questions that require reasoning across multiple, temporally separated video segments.
 VideoChain features a modular architecture built on a modified BART backbone enhanced with video embeddings, capturing textual and visual dependencies.
Using the TVQA+ dataset, we automatically construct the large-scale MVQ-60 dataset by merging zero-hop QA pairs, ensuring scalability and diversity. Evaluations show VideoChain strong performance across standard generation metrics: ROUGE-L (0.6454), ROUGE-1 (0.6854), BLEU-1 (0.6711), BERTScore-F1 (0.7967), and semantic similarity (0.8110).
These results highlight the model’s ability to generate coherent, contextually grounded, and reasoning-intensive questions.
To facilitate future research, we publicly release our code and dataset\footnote{\url{https://github.com/AnupamPandey199949/VideoChain}}. 
\end{abstract}

\section{Introduction}
The field of Question Generation (QG) has garnered substantial attention for its potential to create interactive and informative learning environments, educational tools, and intelligent systems. Traditionally, QG systems focused on generating questions based on textual passages, with applications spanning from educational quizzes \cite{krishna2015automatic} to interview questions and clarification prompts \cite{kumar2020clarq}. 
However, the exploration of QG in video-based content remains significantly underdeveloped compared to its text-based counterpart. 

Video Question Generation (VideoQG) involves generating questions based on the visual and textual information available in video content. This task is particularly important for assessing a model's ability to understand and reason over dynamic and temporal data, as videos often present information across multiple frames, segments, and visual cues. 
Despite its potential, existing VideoQG research has mostly focused on generating zero-hop questions from transcripts or basic visual elements in videos, limiting the scope of reasoning required from the model. In contrast to zero-hop questions, multi-hop question generation requires reasoning across multiple, often non-contiguous, segments of data. Although this has been extensively explored in the text domain through datasets like HotpotQA \cite{yang2018hotpotqa}, it remains relatively underexplored in video-based tasks. Multi-hop questions challenge the model to synthesize information across several frames or segments in a video, thus requiring a more in-depth understanding of both visual and textual modalities.

To address this gap in the literature, we introduce the task of Multi-hop Video Question Generation (MVQG), where the goal is to generate questions that require reasoning over multiple, temporally-separated segments of a video. To facilitate this task, we construct a new MVQG dataset (MVQ-60) by merging zero-hop questions from the TVQA+ dataset \cite{lei-etal-2020-tvqa}. Inspired by the MusiQue paper \cite{trivedi2022musique}, which demonstrated the effectiveness of merging simple questions to create multi-hop questions in the text domain, we extend this concept to videos.
This dataset is designed to challenge models to process and integrate information from different parts of a video, combining both visual frames and textual transcripts. Additionally, we fine-tune the BART \cite{lewis2020bart}, enhanced with video embeddings, to generate coherent, contextually rich multi-hop questions that span multiple video segments.
The fine-tuned models developed in this work demonstrate a significant improvement in generating complex, multi-hop video-based questions
both in terms of automatic evaluation metrics and human judgment.

We summarize the contributions of our work as: 
\textbf{(1)} We contribute the first MVQG system by fine-tuning a customized BART architecture, incorporating text and video embeddings, to generate multi-hop questions requiring reasoning across video segments.
\textbf{(2)} We introduce MVQ-60, the first dataset, specifically designed for MVQG, created by merging zero-hop questions from the TVQA+ dataset, enabling complex reasoning over multiple video segments.

\section{Related Work}
\subsection{Text-based Question Generation}
Question Generation (QG) in text has been extensively studied, with works focusing on various levels of textual granularity. These include document-level QG \cite{pan2020semantic}, paragraph-level QG \cite{zhang2017automatic}, sentence-level QG \cite{ali2010automatic}, and keyword-based QG \cite{pan2020learning}. Early works employed rule-based approaches, but recent advancements leverage deep learning models, particularly sequence-to-sequence architectures and pre-trained transformers \cite{pan2019recent}. Techniques, such as semantic parsing and reinforcement learning have further enhanced the quality of generated questions \cite{chatterjee2020faqaugmenter}. While text-based QG has matured significantly, the shift towards multimodal domains presents new challenges, particularly in reasoning over both visual and temporal modalities.

\subsection{Visual Question Generation}
Visual Question Generation (VQG), introduced by \cite{mostafazadeh2016generating}, generates questions from images and encompasses three types: Visually Grounded (answerable from the image \cite{antol2015vqa}), Commonsense-Based (requiring external knowledge \cite{wang2017fvqa}), and World Knowledge-Based (integrating factual knowledge bases \cite{shah2019kvqa}). Proposed methods include encoder-decoder \cite{mostafazadeh2016generating}, compositional \cite{liu2018ivqa}, and generative models \cite{jain2017creativity}, enhanced by reinforcement learning \cite{yang2018visual} and bilinear pooling \cite{fukui2016multimodal}. Domain-specific applications (e.g., medical imaging, education \cite{mehta2024circuitvqa}) exist, but challenges persist in visual grounding, multi-object reasoning, and extending these challenges to video.

\subsection{Video Question Generation}
VideoQG is inherently more challenging than text or visual QG due to the temporal structure and multimodal nature of videos. Early works \cite{yang2021just} primarily focused on generating questions based on video transcripts or static object and attribute descriptions \cite{gupta2022newskvqa},
but fell short of addressing more complex reasoning requirements.
Multi-hop reasoning in video QG presents unique challenges: Contextual Integration: Generating self-contained questions that require linking temporally distant events in a video.
Entity-Action Mapping: Associating visual entities with their respective actions or interactions in a coherent manner.
Multimodal Fusion: Effectively leveraging signals from various modalities (e.g., video frames, audio, and textual subtitles) to generate questions that reflect comprehensive reasoning.
Existing VideoQG datasets \cite{gupta2022newskvqa, acharya2019tallyqa} target zero-hop question generation, lacking support for reasoning across multiple video segments. While recent video-language models like Flamingo \cite{alayrac2022flamingovisuallanguagemodel} and Vid2Seq \cite{yang2023vid2seq} advance video understanding, they remain limited for multi-hop question generation evaluation.

Our work addresses these gaps by introducing a novel dataset MVQ-60 and developing method specifically designed for multi-hop reasoning over videos
(see table~\ref{tab:approach-comparison}).

\section{Datasets}

VideoQA progress stems from datasets with distinct challenges.
Existing datasets, such as MSR-VTT \cite{7780940} (open-domain video captioning), HowTo100M \cite{miech19howto100m} (instructional videos), TVQA+~\cite{lei-etal-2020-tvqa} (narrative comprehension), ActivityNet-QA \cite{article}(grasping complex videos), and others have been useful in evaluating the performance of VideoQA models. 
While existing datasets focus on zero-hop questions (answerable from single events), multi-hop reasoning across video segments remains underexplored. 

\subsection{Dataset Creation}
Recognizing the lack of multi-hop VideoQA datasets, we opted to create a new dataset. 
Given the scalability and reproducibility challenges of manual annotation, we pursued the development of an automated process to generate multi-hop questions by using existing datasets. This decision was inspired by the MUSIQUE dataset \cite{trivedi2022musique}, which successfully generated textual multi-hop questions through automated merging techniques. Our methodology involved merging two zero-hop video questions to form a multi-hop video question.
Among the 40 datasets reviewed, TVQA+~\cite{lei-etal-2020-tvqa} consisting of 152,545 QA pairs from 21,793 clips, spanning over 460 hours of videos based on 6 popular TV shows(The Big
Bang Theory, How I Met Your Mother, Friends, House M.D., Grey's Anatomy, and Castle), emerged as the optimal base for dataset construction. Its advantages include:

\textbf{High Annotation Quality:} Questions and answers are manually crafted, ensuring accuracy and relevance.
\textbf{Contextual Richness: }Includes subtitles, video frames, and metadata (e.g., episodes, seasons).
\textbf{Widespread Use:} Recognized as a benchmark in VideoQA research, ensuring broad compatibility with existing methods. the TVQA+'s extensive coverage of temporally rich, real-world scenarios provided an ideal foundation for our multi-hop dataset.

\subsection{Automated multi-hop Question Generation}
To generate multi-hop questions, we developed a rule-based merging algorithm inspired by the MUSIQUE dataset’s textual question generation strategy \cite{trivedi2022musique}, that combines pairs of zero-hop questions into coherent multi-hop questions. The key steps in the algorithm are:

\textbf{Question Filtering:} Short questions with concise answers were prioritized to maintain readability and prevent excessive length in merged multi-hop questions. Based on the empirical distribution of the TVQA+ dataset and experimental validation, we set the length thresholds to 15 words for questions and 3 words for answers. These thresholds ensured broad coverage of the dataset while avoiding verbosity, and led to the most effective generation of coherent multi-hop questions.

\textbf{Temporal and Contextual Matching}: The questions were grouped based on the shared metadata, specifically the episode. Let,\noindent $\mathcal{M} =\{m_1, m_2,\ldots, m_k\}$ represent the metadata attributes, where $m$ includes the episode ($e$) and segment ($s$). Two questions $q_i$ and $q_j$ are considered \textbf{Temporally and contextually aligned} if they share the same episode but have different segments:
\[\scriptsize
\text{Match}(q_i, q_j) = \begin{cases} 
1, & \text{if } e_i = e_j \text{ and } s_i \neq s_j \\
0, & \text{otherwise}
\end{cases}
\]
This ensures that only questions referring to the same episode but different segments are considered for merging.

\textbf{Overlap Detection:} Overlap between two questions is defined as instances where the answer to one question forms a semantic part of another question. Let $q_1$ and $q_2$ be two questions with answers $a_1$ and $a_2$, respectively. Overlap is defined as:
\[\scriptsize
\text{Overlap}(q_1, a_2) = \begin{cases} 
1, & \text{if } a_2 \in q_1  \\
0, & \text{otherwise}
\end{cases}
\]

\begin{figure}[t]
    \includegraphics[width=0.9\linewidth]{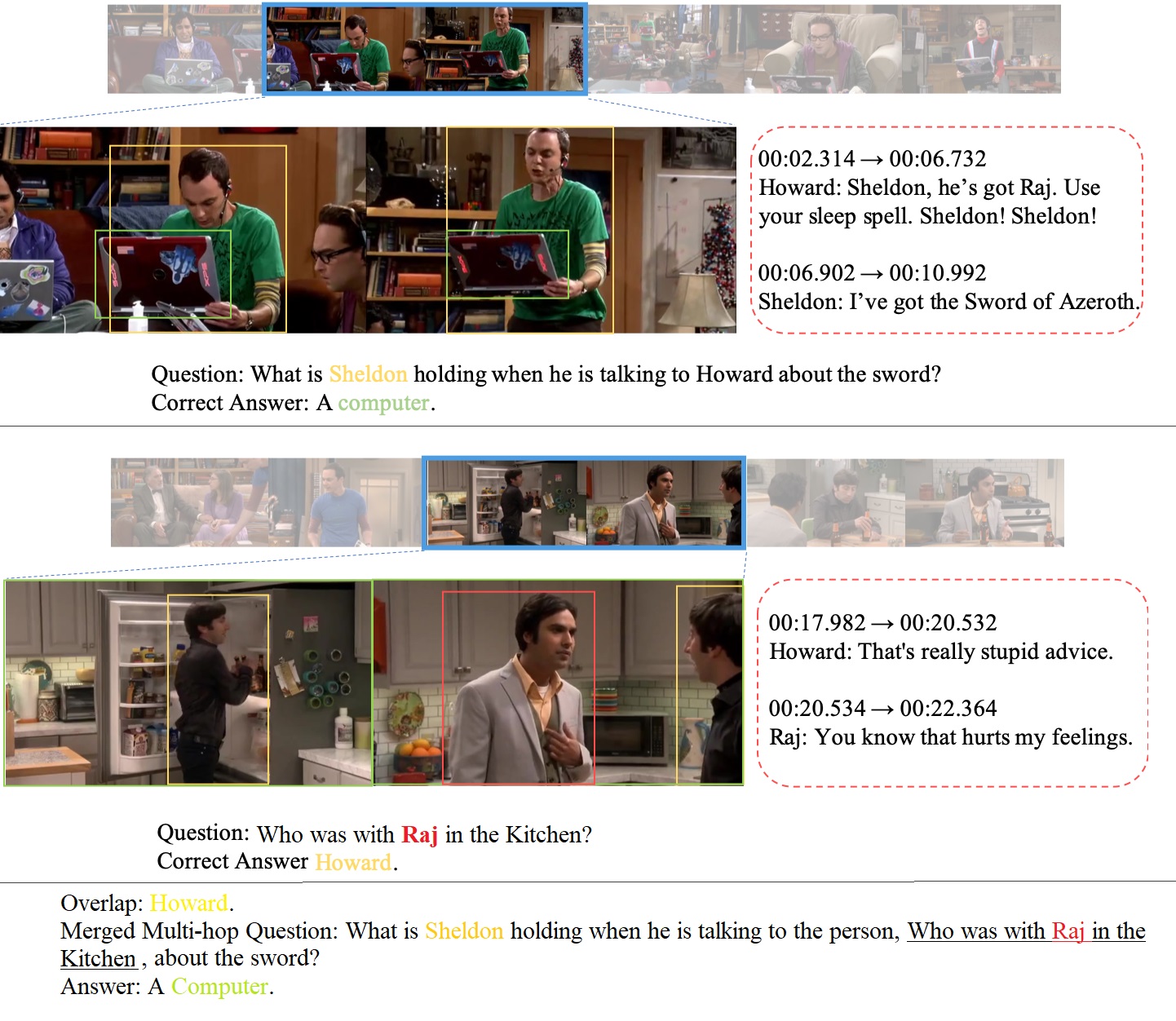} 
    \scriptsize
    \caption{Example: Merged Multi-hop Question}
    \label{fig:Example_Question}
\end{figure}

\textbf{Question Pairing and Merging:} Pairs of overlapping questions were merged to create multi-hop questions by replacing the overlap in first question $q_1$ and second answer $a_2$ with the second question $q_2$. The merged question $q_{\text{merged}}$ is defined as:
\[\scriptsize
q_{\text{merged}} = q_1 \setminus a_2 + q_2
\]
where $q_1 \setminus a_2$ denotes the replacement of $a_2$ in $q_1$ with $q_2$'s context.
An example of this process is shown in Figure \ref{fig:Example_Question}.
\subsection{Quality Evaluation Metrics}
To assess the quality of the generated questions, we evaluate them using a range of metrics:
\textbf{Fluency:} Grammatical correctness and natural language quality.
\textbf{Multi-Hop Reasoning:} Complexity of reasoning required to answer the question.
\textbf{Video Relevance:} Degree of relevance to one or both videos.
\textbf{Engagingness:} How captivating and interesting the question is.
\textbf{Factual Correctness:} Logical and factual accuracy of the question-answer pairs.
\textbf{Inclusiveness:} How well the questions covered diverse aspects of the video content.
Each metric was scored on a scale of 0 to 3, where 3 indicated the highest quality. For example:

Fluency: ``What is Chandler's wife cooking?'' → Score: 3 (Excellent)
Relevance: ``Compare the styles of dance by people in both videos.'' → Score: 3 (Highly Relevant)

For scalability, we conducted evaluation on a random sample of 200 questions. According to statistical sampling theory, such a subset can provide reliable estimates of overall dataset characteristics.
To further validate the dataset’s quality and ensure its appropriateness for multi-hop video QA tasks, we performed manual human evaluation on the sampled set. Annotators scored each question across the above six criteria using predefined rubrics. The average scores were: Fluency(2.92), Multi-hop Reasoning(3.00), Engagingness(2.80), Factual Correctness(3.00).
The high scores, particularly in reasoning and factual correctness, are attributed in part to the use of high-quality human annotated TVQA base questions, which were merged to construct MVQ-60.
All annotations were done by three trained annotators following strict rubrics which ensured high objectivity and achieved an inter-rater agreement (Cohen’s~$\upkappa$) of \textbf{ 0.72}. These results reinforce the validity and challenge-level of MVQ-60 for future multi-hop video reasoning research.
Finally, we introduce MVQ60, the first large-scale dataset of multi-hop (two-hop) video questions, consisting of over 60,000 questions based on six popular TV shows (Friends, The Big Bang Theory, How I Met Your Mother, House M.D., Grey's Anatomy, Castle), with an average question length of ~27 words.

\section{Methodology}

\noindent\textbf{Video Embedding Generation:}
A key aspect of our methodology was the generation of expressive video embeddings to encode spatio-temporal dynamics. We used VideoMAE, a masked autoencoder for self-supervised video representation learning \cite{10203656}. VideoMAE embeddings effectively capture both motion and appearance features while maintaining computational efficiency. These embeddings served as the foundational representation for all the subsequent experiments and models.

\noindent\textbf{Initial Exploration with Video-Based Models:}
We began by finetuning state-of-the-art video-based models, pretrained on tasks, such as video captioning and VideoQA, to adapt them for MVQG. These models were trained using VideoMAE embeddings and textual inputs (e.g., video transcripts). This approach was inspired by works like \cite{lei-etal-2020-tvqa}, which emphasized the use of spatio-temporal features for question answering, and \cite{article}, which demonstrated the benefits of video pretraining for understanding complex narratives.
While these models showed promise in zero-hop reasoning tasks, they struggled with multi-hop questions. Their large sizes and monolithic architectures resulted in slow processing times and difficulty scaling to higher-hop reasoning. These limitations prompted us to explore lightweight, text-based alternatives.

\noindent\textbf{Transition to Text-Based Models:}
Inspired by \cite{phukan-etal-2024-ecis} that text-based models can generate high-quality video questions with video embeddings, we explored using T5 \cite{2020t5}, BART \cite{lewis2020bart}, and similar models. 
We modified these text-based models to accept VideoMAE embeddings as additional input alongside textual data. During finetuning, the models were provided with both embeddings and transcripts. While this setup improved efficiency, the generated questions occasionally failed to integrate information across video segments coherently. These challenges, highlighted the need for explicit architectural modifications to handle multi-hop reasoning.
\noindent\textbf{Modular Two-Component Architecture:}
To address the observed limitations, we develop a model with two-component architecture tailored for multi-hop question generation. This modular design was inspired by the principles of \cite{andreas2016neural}, which demonstrated the benefits of task decomposition for reasoning tasks, and \cite{trivedi2022musique}, which emphasized modular frameworks for multistep reasoning.
The first component generates zero-hop questions from individual video segments. It accepts as input the VideoMAE embedding of a video clip, its corresponding transcript, and a prompt that guides the question generation process. The model outputs a concise, contextually grounded question. For example, given a clip where Monica is cooking, this component generates the question “What is Monica cooking?”. This step isolates relevant information from each video segment, forming a foundation for multi-hop reasoning.
\begin{figure*}[t]:
    \centering
    \includegraphics[width=0.9\textwidth]{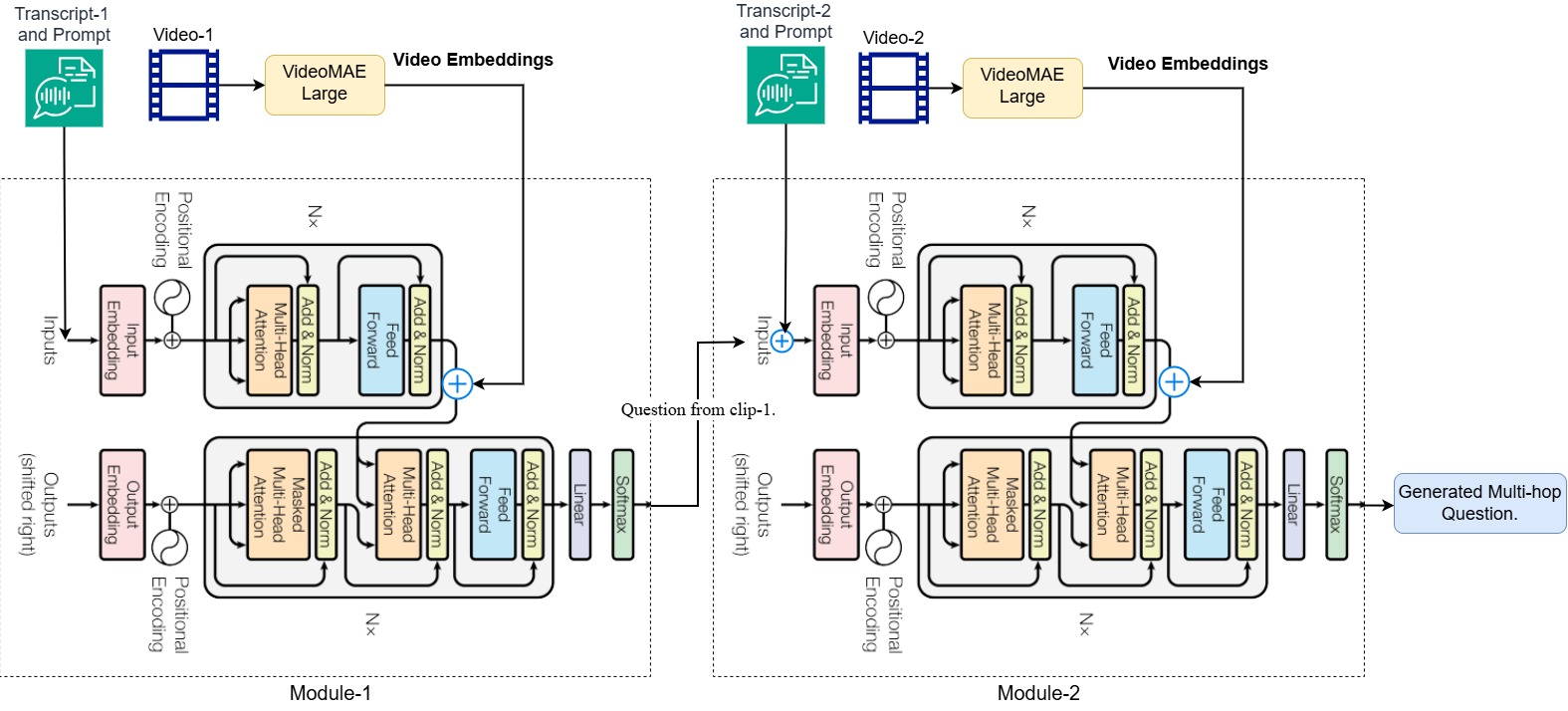} 
    \scriptsize
    \caption{Proposed Model Architecture}
    \label{fig:model_architecture}
\end{figure*}
The second component integrates information from multiple video segments to generate multi-hop questions. It accepts as input the zero-hop question produced by the first component, along with the VideoMAE embedding and transcript of a second video segment. The model refines and expands the question, producing a multi-hop question, such as “What is Chandler’s wife cooking?”. This process can be repeated iteratively, allowing the model to generate higher-hop questions (e.g., three-hop or four-hop questions) by forwarding the output question to subsequent iterations with new inputs.


\noindent\textbf{Two-Component Approach:} The modular nature of the two-component architecture offers several key benefits. By decoupling zero-hop and multi-hop reasoning, the system scales efficiently to higher-hop questions without overwhelming model capacity. Each component specializes in a specific task, improving both precision and contextual grounding. 

\section{Model Architecture}
The proposed architecture for MVQG (figure \ref{fig:model_architecture}) centers around VideoChain, a version of the BART-large CNN model \cite{lewis2020bart} we modified to process both video and textual inputs as distinct modalities. It integrates the spatio-temporal dynamics of video content through VideoMAE embeddings \cite{10203656} while preserving BART’s generative capabilities for text-based reasoning. 
VideoChain processes two primary input types: video embeddings and text embeddings. Video embeddings are generated using VideoMAE, a self-supervised video representation learning model that encodes video segments into $\mathbb{R}^{1568 \times 1024}$ dimensional feature vectors. These embeddings capture both motion and appearance features, providing a compact representation of the video’s spatio-temporal content. 
Text embeddings are derived from video transcripts and prompts.
Notably, the prompts differ between components; the first component uses prompts to generate zero-hop questions (e.g., “Generate a question about this clip”), while the second component uses prompts to guide the generation of multi-hop questions (e.g., “Generate a multi-hop question based on the previous question and this clip”).
The architecture builds upon BART-large CNN by introducing modifications that enable it to handle multimodal inputs effectively. First, the encoder is extended to include dual input streams for video and text embeddings. The video embeddings are processed through dedicated multi-head attention and feedforward layers designed for spatio-temporal data, while the text embeddings are processed through the standard BART encoder layers. A cross-modal attention mechanism is introduced to fuse the outputs of the video and text streams, enabling VideoChain to reason jointly over both modalities. The decoder attends to the fused multimodal representation, generating the output question token by token. We use this VideoChain in both modules of our architecture.

\noindent\textbf{Module 1:Zero-hop Question Generation:}
The zero-hop question generation component constitutes the first stage of the architecture. This component generates a concise question based on the content of a single video segment. The video embeddings, transcripts, and prompts are processed independently through their respective streams in VideoChain’s encoder. The cross-modal attention mechanism aligns the visual and textual representations, producing a unified multimodal encoding that informs the decoder’s generation process. For example, given a video clip where Monica is cooking, the model generates the question “What is Monica cooking?”. The training process for this component employs cross-entropy loss. By focusing on zero-hop reasoning, this component ensures that VideoChain can effectively extract and represent information from individual video segments.

\noindent\textbf{Module 2:Multi-hop Question Composition:}
The second stage of the architecture, the composition of multi-hop questions, extends the zero-hop question by incorporating additional information from subsequent video segments. This component takes as input the zero-hop question generated by the first component, along with the video embedding, transcript, and a multi-hop-specific prompt corresponding to the second video segment. The encoder processes these inputs in their respective streams, and the cross-modal attention mechanism aligns the zero-hop question with the visual and textual context of the second segment. The decoder refines and expands the zero-hop question into a multi-hop question. For instance, if the zero-hop question is “What is Monica cooking?” and the second segment provides context about Monica’s relationship with Chandler, Module-2 generates the multi-hop question “What is Chandler’s wife cooking?”. This iterative design enables the architecture to handle reasoning tasks of arbitrary complexity by recursively invoking the second component with new inputs.

\noindent\textbf{Training Strategy:}
Our model is trained in two stages to optimize its performance for zero-hop and multi-hop tasks. The zero-hop Question Generation component is trained on zero-hop question-answer pairs using cross-entropy loss, while the multi-hop Question Composition component is trained on multi-hop question-answer pairs, with ground-truth intermediate questions provided during training. A composite loss function is used for the second component, combining cross-entropy loss with alignment loss to ensure effective multimodal fusion.
During inference, our model uses beam search to enhance the fluency and coherence of the generated questions, particularly for higher-hop reasoning tasks where maintaining logical consistency is critical.
\noindent\textbf{Scalability and Adaptability:}
Our model’s modularity ensures scalability by isolating reasoning subtasks into distinct components. The recursive nature of the multi-hop component allows the system to handle increasingly complex tasks without requiring additional architectural changes. 
VideoChain’s flexibility allows it to integrate with other pre-trained generative models. While BART-large CNN serves as the base in this implementation, the modifications applied to handle video and text inputs can be extended to models, such as T5 \cite{2020t5} or mBART \cite{liu2020multilingualdenoisingpretrainingneural}, enabling the framework to adapt to a wide range of datasets and applications. This adaptability is crucial for advancing multimodal reasoning in 
VideoQA \cite{zellers2019vcr}.




\section{Experiments, Results and Analysis}

\subsection{Experiment Setup}

For our experiments, we used the MVQ-60 dataset,
which comprises 60,000 multi-hop questions paired with video segments and corresponding transcripts. The dataset was split into 80\% training, 10\% validation, and 10\% test, ensuring no episode level overlap between these splits to prevent data leakage and overfitting. During the fine-tuning step, we used the input IDs and attention masks of the inputs, which consisted of concatenated prompts and transcripts representing the questions, and passed the video embeddings as a separate entity into our model. The model was trained on 2 Tesla T4 GPUs on kaggle for a total of 8 hours.
\textit{Hyperparameters:} a learning rate of 3e-5, a batch size (4)suitable for the available hardware, and gradient accumulation steps set to 4. Training was conducted for 50 epochs. We used a mixed precision training approach with FP16 enabled for compatibility with CUDA, and we monitored performance using the evaluation strategy set to run every 100 steps.
The maximum gradient norm was clipped to 1.0 to ensure stable training, while unnecessary columns were removed to optimize memory usage. 
Despite being a compact architecture (~406M parameters, using BART-large), our model delivers strong multi-hop reasoning performance with significantly lower training time and resource requirements compared to large-scale vision-language models. For example, Qwen2-VL-2B requires approximately 28 hours of training on similar data, whereas our VideoChain-based model achieves competitive results with just 8 hours of training, highlighting the efficiency and scalability of our approach.

\subsection{Experiments}

\begin{table*}[t!]
    \centering 
    \scriptsize
    \caption{Results on Human Evaluation}
    \begin{tabular}{lcccccc}
        \hline
        \textbf{Model} & \textbf{Fluency} & \textbf{Relevance} & \textbf{Multi-Hop Reasoning} & \textbf{Engagingness}&  \textbf{Factual Correctness} & \textbf{Inclusiveness}\\
        \hline
        \textbf{VideoChain (Ours)} & 2.89 & \textbf{2.91} & \textbf{2.81} &\textbf{ 2.75} & \textbf{2.92 }& \textbf{2.78}\\
        ECIS (Finetuned) & 2.90 & 2.85 & 2.72 & 2.63 & 2.84 & 2.65\\
        Qwen2-VL-2B-Instruct & 2.71 & 2.32 & 1.54 & 2.25 & 2.56 & 2.05 \\
        SmolVLM & 2.51 & 2.09 & 1.24 & 1.82 & 1.98 & 2.24\\
        MGM-2B & 2.80 & 2.54 & 1.64 & 2.21 & 2.24& 1.97\\
        PaliGemma & \textbf{2.95} & 2.60 & 1.77 & 2.41 & 2.18 & 2.14\\
        \hline
    \end{tabular}
    \label{tab:human_eval}
\end{table*}

\begin{table*}[t!]
    \centering
    \scriptsize
    \caption{Results on Automatic Evaluation}
    \begin{tabular}{lcccccccc}
        \hline
        \textbf{Model} & \textbf{Bert score F1} & \textbf{Generation length} & \textbf{Semantic similarity} & \textbf{Rouge-1} & \textbf{Rouge-L} & \textbf{Bleu-1} & \textbf{Distinct-1} & \textbf{Distinct-2}\\
        \hline
        \textbf{VideoChain (Ours)} & \textbf{0.7967} & \textbf{53.2} & \textbf{0.8110} & \textbf{0.6854} & \textbf{0.6454} & \textbf{0.6711} & \textbf{0.7911} & \textbf{0.9850}\\
        ECIS (Finetuned) & 0.6253 & 47.5 & 0.5291 & 0.4006 & 0.3174 & 0.4203 & 0.7430 & 0.9553\\
        Qwen2-VL-2B-Instruct & 0.5120 & 75.3 & 0.4547 & 0.2746 & 0.2451 & 0.3712 & 0.7230 & 0.9370 \\
        SmolVLM & 0.4881 & 78.1 & 0.5154 & 0.2819 & 0.2482 & 0.3574 & 0.7115 & 0.9260 \\
        MGM-2B & 0.5046 & 70.7 & 0.5627 & 0.3004 & 0.2627 & 0.3861 & 0.7250 & 0.9410\\
        PaliGemma & 0.5287 & 82.6 & 0.4987 & 0.2912 & 0.2519 & 0.4021 & 0.7340 & 0.9505\\
        \hline
    \end{tabular}
    \label{tab:Auto_eval}
\end{table*}
As VideoChain represents the first dedicated architecture for MVQG, there are no prior models explicitly trained for this task. To establish meaningful baselines, we conducted zero-shot evaluations using recent general-purpose multimodal models: Qwen2-VL-2B-Instruct, SmolVLM, MGM-2B, and PaliGemma. These models were selected for their ability to process video and language inputs without task-specific fine-tuning. In addition, to assess the adaptability of existing VideoQA models, we finetuned the ECIS model \cite{phukan-etal-2024-ecis} on our MVQ-60 dataset. This provides a strong baseline from the VideoQA domain.
This evaluation highlights the limitations of generic vision-language models and repurposed QA models in generating coherent, compositional video-grounded questions.

\subsection{Evaluation Setup}\label{sec:evaluation-setup}
For Human Evaluation: To ensure a robust and diverse evaluation, we recruited human annotators from various demographic backgrounds. Our annotators consisted of individuals proficient in English, with a mix of undergraduate and graduate students.
Each evaluator was tasked with assessing a randomly sampled subset of multi-hop questions based on predefined quality metrics. To maintain fairness and ethical considerations, all the annotators were compensated at a competitive rate in accordance with the standard research compensation guidelines.

For zero-shot evaluation, we provided each model with pairs of video segments, corresponding transcripts, and prompts designed to elicit multi-hop reasoning. The prompts were standardized across models to ensure fair comparison. For example: \textit{``Based on the two video segments and their transcripts, generate a question that requires integrating information from both videos.''}
The models' outputs were evaluated on both automated and human evaluation metrics, including \textit{fluency}, \textit{relevance}, \textit{multi-hop reasoning}, \textit{factual correctness}, \textit{engagingness} and \textit{inclusiveness}.
\subsection{Results Analysis}
  As summarized in Tables~\ref{tab:human_eval} and~\ref{tab:Auto_eval}, our proposed \textbf{VideoChain} model consistently outperforms recent multimodal baselines across both human and automatic evaluation metrics.
In \textbf{human evaluation}, while all models demonstrate strong \textbf{fluency} and \textbf{engagingness} due to their pretrained language modeling capabilities,\textbf{VideoChain} achieves the highest scores in \textbf{relevance (2.91)}, \textbf{multi-hop reasoning (2.81)}, and \textbf{factual correctness (2.92)}. The fine-tuned ECIS model also shows strong performance, particularly in fluency (2.90) and correctness (2.84), but falls slightly short in relevance and multi-hop depth. The relatively low factual correctness scores of models, such as Qwen2-VL (2.56), PaliGemma (2.20), and SmolVLM (1.98) suggest hallucination and external knowledge leakage, likely caused by exposure to the same TV shows during pretraining. In contrast, VideoChain is explicitly trained on grounded supervision, encouraging content-aligned generation.
The \textbf{inclusiveness} metric further highlights the strength of our architecture. As most baselines treat both video segments jointly, they often produce questions grounded in only one clip. VideoChain’s modular dual-stage design processes each clip in separate stages, enabling more inclusive question generation (\textbf{2.78} compared to 2.05 for Qwen2-VL and 2.14 for PaliGemma).
For \textbf{multi-hop reasoning}, baseline models frequently concatenate multiple independent zero-hop questions rather than forming a coherent multi-hop question. Our model, trained with explicit multi-hop supervision, demonstrates better temporal and semantic integration across clips.
In \textbf{automatic evaluation}, VideoChain leads across all metrics, BERTScore F1 (0.7967), semantic similarity (0.8110), ROUGE-1 (0.6854), ROUGE-L (0.6454), BLEU-1 (0.6711), Distinct-1 (0.7911), and Distinct-2 (0.9850). These gains show that VideoChain generates fluent, diverse, and semantically aligned multi-hop questions. ECIS also shows competitive performance, especially in fluency-aligned metrics, confirming the benefits of fine-tuning on MVQ-60. while the other Baseline models often generate overly long questions (e.g., PaliGemma: 82.6 tokens, SmolVLM: 78.1), which negatively impacts lexical and semantic alignment, the stricter length-controlled finetuning of VideoChain and ECIS, helped them produce more concise and accurate questions.

\subsection{Discussion}

The zero-shot evaluation highlights the limitations of current pre-trained models in addressing multi-hop VideoQA tasks without specialized training. While models like Qwen2-VL-2B-Instruct and PaliGemma show promise in handling general vision-language tasks, their performance on multi-hop reasoning remains suboptimal compared to our fine-tuned model. This suggests that while general-purpose multimodal models offer flexibility, task-specific architectures like VideoChain are essential to achieve state-of-the-art performance in complex reasoning tasks, such as MVQG.

\subsection{Ablation Study} We assess the contribution of VideoChain's core components through two ablations:
(1) Text-only variant: Removing video embeddings to isolate text reliance caused significant performance drops, particularly in video relevance (2.91 to 2.09) and multi-hop reasoning (2.85 to 1.54). This confirms visual grounding is essential for contextually rich question generation.
(2) Single-component variant: Replacing the modular pipeline with direct multi-hop generation severely degraded reasoning capability (multi-hop: 1.24 vs. 2.85) and factual correctness (1.98 vs. 2.97), often yielding shallow or concatenated questions.
Both variants showed substantial overall performance degradation (0.69 text-only; 0.76 single-component) versus the full model. These results validate the necessity of multimodal inputs for visual grounding and modular decomposition for complex reasoning. Detailed metrics and analysis are provided in the appendix section \ref{app: Ablation Study}.

\subsection{Error Analysis}
To better illustrate the strengths and limitations of our MVQG model, we present qualitative examples across the six TVQA+ shows in Table~\ref{tab:mvqg_errors}. Each row provides one example from a distinct TV show, categorized into four groups: (1) \textbf{Correct Generations}, (2) \textbf{Multi-Hop Reasoning Failures}, (3) \textbf{External Knowledge Leakage}, and (4) \textbf{Hallucination}. Detailed Error Analysis is discussed in the Appendix section \ref{app: error_analysis}

\section{Conclusion and Future Work}
We introduced VideoChain, the first modular architecture for MVQG. We created MVQ-60, a large-scale multihop video question dataset spanning six TV shows. VideoChain’s modular design ensures scalability for complex reasoning. Evaluations demonstrated its strong performance in generating fluent, relevant, and coherent video-grounded multihop questions, validating our approach. We evaluated the intrinsic quality of the generated questions (using automatic metrics and human judgements), but we did not perform an extrinsic evaluation, e.g., using the questions for a downstream task like training a VideoQA model or testing a model’s reasoning capabilities. 
Future work includes expanding to diverse domains (e.g., education, surveillance), developing domain-specific MVQG systems, enabling multilingual generation, and integrating emerging vision-language models for enhanced reasoning and nuance.
\section{Limitations}
Despite the advancements demonstrated in this work, several limitations warrant further investigation. Expanding to diverse video domains (e.g., educational, surveillance) requires tackling distinct features and QA demands, likely needing domain-specific adaptations. Second, our system generates questions only in English. Enabling multilingual question generation is crucial for broader accessibility but involves complex cross-lingual understanding and generation. Third, the modular pipeline risks error propagation; factual errors from Module-1 often persist despite Module-2's mitigation of grammatical/semantic issues mitigation. While not significantly impacting overall quality, explicit error correction or joint optimization could help. Finally, rapidly evolving vision-language models offer potential for more powerful representations, reasoning, and nuanced question generation. Future work should integrate these to achieve deeper video understanding. Furthermore, since our work does not explicitly leverage raw audio signals or audio features (beyond what is indirectly available through dialogue text), future work could include tri-modal input (video, text, audio) for better grounding. Additionally, explicit error correction or a joint optimization of both modules could be explored to mitigate the propagation of factual errors from Module-1 to Module-2.

\bibliography{custom}

\appendix
\section{Appendix}
\section{Automated MVQ-60 Dataset Generation}
We propose a rule-based algorithm for generating multi-hop questions by merging zero-hop question-answer pairs, inspired by MUSIQUE \cite{trivedi2022musique}:

\begin{algorithm}[H]
\caption{MVQ-60 Generation}
\label{alg:multihop}
\begin{algorithmic}[1]
\REQUIRE Set of zero-hop QA pairs $\{(q_i, a_i, m_i)\}$ with metadata $m_i = (e_i, s_i)$
\ENSURE Set of multi-hop questions $\mathcal{Q}_{\text{multi}}$

\STATE Filter QA pairs: $\mathcal{Q}_{\text{filtered}} \gets \{(q, a) \mid \text{len}(q) \leq 15,  \text{len}(a) \leq 3\}$
\FOR{each episode $e$}
    \STATE Group pairs: $\mathcal{G}_e \gets \{(q_i, a_i, s_i) \mid e_i = e\}$
    \FOR{each pair $(q_1, a_1, s_1), (q_2, a_2, s_2) \in \mathcal{G}_e \times \mathcal{G}_e$}
        \IF{$s_1 \neq s_2$ \AND $a_2$ is substring of $q_1$}
            \STATE $q_{\text{merged}} \gets \text{replace}(q_1, a_2, q_2)$
            \STATE $\mathcal{Q}_{\text{multi}} \gets \mathcal{Q}_{\text{multi}} \cup \{q_{\text{merged}}\}$
        \ENDIF
    \ENDFOR
\ENDFOR
\end{algorithmic}
\end{algorithm}

The algorithm processes filtered QA pairs (len(q) $\leq$15, len(a) $\leq$3) grouped by episode. For valid pairs from different segments where $a_2$ appears in $q_1$, it replaces $a_2$ in $q_1$ with $q_2$ to form multi-hop questions requiring cross-segment reasoning.

\section{Models Evaluated}

We evaluate the following pretrained models for MVQG:

     \textbf{Qwen2-VL-2B-Instruct} \cite{Qwen2VL}: A vision-language model optimized for instruction-following tasks, capable of integrating visual inputs with complex text prompts. It was tested using raw video frames and associated transcripts.
    
    \textbf{SmolVLM} \cite{marafioti2025smolvlm}: A lightweight multimodal large language model designed for efficient inference on vision-language tasks. Despite its compact size, SmolVLM demonstrated strong performance on zero-hop VQA tasks, but its ability to handle multi-hop reasoning remained untested prior to our evaluation.
    
    \textbf{MGM-2B} \cite{li2024mgm}: A 2-billion parameter multimodal generative model designed for cross-modal understanding and generation tasks. It processes video frame embeddings and textual data simultaneously, offering a comprehensive baseline for multimodal reasoning.
    
    \textbf{PaliGemma} \cite{beyer2024paligemmaversatile3bvlm}: A recent multimodal model designed by Google DeepMind for VQA and vision-language reasoning tasks. Although It excels in vision-language alignment, its capacity for multi-hop reasoning with video content was tested in our experiments.
    
    \textbf{ECIS-VQG} \cite{phukan-etal-2024-ecis}: A VideoQG model, which was designed to produce entity-centric, information-seeking questions grounded in video content. Originally proposed for Zero-hop VideoQA tasks, we include ECIS-VQG to evaluate its capacity for generalizing to the multi-hop setting when finetuned on our MVQ-60 dataset.

\begin{table}[t]
\scriptsize
\caption{Comparison of Different Approaches}
\label{tab:approach-comparison}
\begin{tabular}{|c|c|c|c|c|}
\hline
Approach & Text & Video & Multihop Reasoning & Video Relevance \\
\hline
Text QG & \checkmark & \texttimes & \checkmark & \texttimes \\
\hline
Video QG & \checkmark & \checkmark & \texttimes & \checkmark \\
\hline
Zeroshot & \checkmark & \checkmark & \checkmark & \texttimes \\
\hline
Our Approach & \checkmark & \checkmark & \checkmark & \checkmark \\
\hline
\end{tabular}
\end{table}
\section{Datasets Explored}

To contextualize the need for a dedicated multihop video question answering (MVQG) dataset and to inform our design choices, we conducted a comprehensive survey of existing video question answering (VideoQA) datasets. This exploration encompassed approximately 40 publicly available datasets, each with varying characteristics in terms of scale, domain, question type, and associated annotations. A summary of these datasets, including their approximate size, primary focus, and question types, is provided in Table~\ref{tab:dataset_summary}.

While these datasets have significantly advanced the field of VideoQA, they primarily focus on zero-hop questions that can be answered by directly attending to specific segments or elements within a single video clip. As highlighted in \textbf{Table~\ref{tab:dataset_summary}}, these datasets cover a diverse range of domains.

Our analysis of these existing resources revealed a critical gap: the absence of datasets specifically designed to evaluate and drive research in multihop video question answering. As detailed in the main body, multihop questions require reasoning across multiple temporal segments or understanding the relationships between different events or entities within and potentially across video clips. The existing datasets, while valuable for zero-hop VideoQA, do not adequately support the investigation of these more complex reasoning capabilities.

To address this limitation and facilitate research into MVQG, we undertook the creation of a novel dataset, leveraging the TVQA+ dataset as a foundation, as described in Section:dataset-creation in main paper. Our approach to automatically generating multihop questions aimed to create a scalable resource for evaluating models capable of performing temporal and relational reasoning across video content.

\section*{Dataset Evaluation}
We prompted GPT5 to evaluate the dataset question quality. Due to cost considerations, we evaluated the same 200-question sample with 300 additional questions.

\noindent \textbf{Prompt: }You are an evaluator. Your task is to assess a generated question based on the context and the following criteria:

\textbf{Fluency:} Evaluates the grammatical correctness and naturalness of the generated question.
     0: Poor (Grammatical errors and awkward phrasing). Example: ``What doing is Chandler wife cooking?''
     1: Fair (Some grammatical errors, but understandable). Example: ``What Chandler wife cooking?''
     2: Good (Grammatically correct). Example: ``What is the wife of Chandler cooking?''
     3: Excellent (Fluent and natural language with no errors). Example: ``What is Chandler’s wife cooking?''

\textbf{Relevance:} Assesses the extent to which the generated question pertains to the content of the provided video clips.
     0: Irrelevant (Does not relate to the video). Example: ``What is the capital of France?''
     1: Slightly relevant (Partially relates to the video). Example: ``Are there any people in the video?''
     2: Mostly relevant (Mostly relates to the Videos). Example: ``Are there people dancing in these videos?''
     3: Highly relevant (Directly relates to the images). Example: ``What is the connection between the people dancing in these Videos?''

\textbf{Multi-Hop Reasoning:} Evaluates the complexity of reasoning required to answer the generated question based on the provided video clips.
     0: Single-hop (Only needs one Video for the answer). Example: ``Is Monica dancing in the first video?''
     1: Simple multi-hop (Requires basic information from both videos). Example: ``Are People dancing in both Videos?''
     2: Intermediate multi-hop (Requires more complex connections between images). Example: ``Are the same people dancing in both videos?''
     3: Advanced multi-hop (Involves detailed reasoning using both images). Example: ``what is the relation between the common people dancing in both videos?''

\textbf{Engagingness:} Evaluates how interesting and captivating the generated question is to a human observer.
     0: Not engaging (Boring or uninteresting). Example: ``Are there hats in the videos?''
     1: Slightly engaging (Mildly interesting). Example: ``What colours are the hats in the videos?''
     2: Moderately engaging (Interesting and engaging). Example: ``How do the styles of hats in the videos differ?''
     3: Highly engaging (Very interesting and captivating). Example: ``What do the hats in the videos reveal about the event going on and time period of the scenes depicted?''

\textbf{Factual Correctness: }Evaluates whether the generated question contains any factual inaccuracies.
     0: Incorrect (factually incorrect). Example: ``Why does Chandler want to leave after hanging out with the group, which includes Amy and Emma?'' (Emma and Amy both are wrong).
     1: Mostly Incorrect (contains major factual errors, though a small part of the content may be accurate). Example: ``Why does Chandler want to leave after hanging out with the group, which includes Joey and Amy?'' (Joey is correct, Amy is wrong).
      2: Partially Correct (factually accurate in the main aspect but contains a minor mistake or omission). Example: ``Why does Chandler want to leave after hanging out with the group, which includes Joey?''
     3: Factually correct. Example: ``Why does Chandler want to leave after hanging out with the group, which includes Joey and Monica?''

\textbf{Inclusiveness:} Evaluates whether the generated question is inclusive and avoids any potentially biased or discriminatory language.
     0: Not inclusive (The question contains biased or discriminatory language or assumptions). Example: ``Why are the women in the video acting emotionally?''
     1: Slightly inclusive (The question is mostly neutral but could be phrased more inclusively). Example: ``What are the people in the video doing?'' (If the context strongly implies a specific gender)
     2: Moderately inclusive (The question attempts to use neutral language but might still have some underlying assumptions). Example: ``What is the role of each person in the scene?''
     3: Highly inclusive (The question uses neutral and respectful language, avoiding any biased or discriminatory assumptions about gender, race, age, etc.). Example: ``What actions are the individuals performing in the video?''

\textit{**Question** **Context**}
\textit{Output Structure: Fluency: value,}
\textit{Relevance: value,}
\textit{Multi-Hop Reasoning: value, Engagingness: value, }
\textit{Factual Correctness: value, }
\textit{Inclusiveness: value,}

\begin{table*}[t!]
    \centering 
    \scriptsize
    \caption{Human and GPT Evaluation the Dataset, MVQ-60 }
    \begin{tabular}{lcccccc}
        \hline
        \textbf{Method } & \textbf{Fluency} & \textbf{Relevance} & \textbf{Multi-Hop Reasoning} & \textbf{Engagingness}&  \textbf{Factual Correctness} & \textbf{Inclusiveness}\\
        \hline
        Human Eval & \textbf{2.92} & \textbf{3} & \textbf{3} &\textbf{ 2.8} & \textbf{3}& \textbf{3}\\
        gpt-5-nano-2025-08-07 & 2.88 & 3 & 2.82 &  2.66 & 2.74 & 2.96\\
       
        \hline
    \end{tabular}
    \label{tab:dataset_eval}
\end{table*}

\section*{Human Evaluation Metrics}

To assess the quality of the generated multihop video questions, we employed a comprehensive human evaluation protocol using the following set of metrics. Each metric was evaluated on a 4-point scale(0 to 3), with higher scores indicating better quality according to the specific criterion.

\textbf{Fluency:} Evaluates the grammatical correctness and naturalness of the generated question.
\begin{itemize}
    \item \textbf{0}: Poor (Grammatical errors and awkward phrasing). \textit{Example: ``What doing is Chandler wife cooking?''}
    \item \textbf{1}: Fair (Some grammatical errors but understandable). \textit{Example: ``What Chandler wife cooking?''}
    \item \textbf{2}: Good (Grammatically correct). \textit{Example: ``What is the wife of Chandler cooking?''}
    \item \textbf{3}: Excellent (Fluent and natural language with no errors). \textit{Example: ``What is Chandler’s wife cooking?''}
\end{itemize}

\textbf{Relevance:} Assesses the extent to which the generated question pertains to the content of the provided video clips.
\begin{itemize}
    \item \textbf{0}: Irrelevant (Does not relate to the video). \textit{Example: ``What is the capital of France?''}
    \item \textbf{1}: Slightly relevant (Partially relates to the video). \textit{Example: ``Are there any people in the video?''}
    \item \textbf{2}: Mostly relevant (Mostly relates to the Videos). \textit{Example: ``Are there people dancing in these videos?''}
    \item \textbf{3}: Highly relevant (Directly relates to the images). \textit{Example: ``What is the connection between the people dancing in these Videos?''}
\end{itemize}

\textbf{Multi-Hop Reasoning:} Evaluates the complexity of reasoning required to answer the generated question based on the provided video clips.
\begin{itemize}
    \item \textbf{0}: Single-hop (Only needs one Video for the answer). \textit{Example: ``Is Monica dancing in the first video?''}
    \item \textbf{1}: Simple multi-hop (Requires basic information from both videos). \textit{Example: ``Are People dancing in both Videos?''}
    \item \textbf{2}: Intermediate multi-hop (Requires more complex connections between images). \textit{Example: ``Are the same people dancing in both videos?''}
    \item \textbf{3}: Advanced multi-hop (Involves detailed reasoning using both images). \textit{Example: ``what is the relation between the common people dancing in both videos?''}
\end{itemize}

\textbf{Engagingness:} Evaluates how interesting and captivating the generated question is to a human observer.
\begin{itemize}
    \item \textbf{0}: Not engaging (Boring or uninteresting). \textit{Example: ``Are there hats in the videos?''}
    \item \textbf{1}: Slightly engaging (Mildly interesting). \textit{Example: ``What colours are the hats in the videos?''}
    \item \textbf{2}: Moderately engaging (Interesting and engaging). \textit{Example: ``How do the styles of hats in the videos differ?''}
    \item \textbf{3}: Highly engaging (Very interesting and captivating). \textit{Example: ``What do the hats in the videos reveal about the event going on and time period of the scenes depicted?''}
\end{itemize}

\textbf{Factual Correctness:} Evaluates whether the generated question contains any factual inaccuracies.
\begin{itemize}
    \item \textbf{0}: Incorrect (factually incorrect). \textit{Example: ``Why does Chandler want to leave after hanging out with the group, which includes Amy and Emma?''}
    \item \textbf{3}: Factually correct. \textit{Example: ``Why does Chandler want to leave after hanging out with the group, which includes joey and monica?''}
\end{itemize}

\textbf{Inclusiveness:} Evaluates whether the generated question is inclusive and avoids any potentially biased or discriminatory language.
\begin{itemize}
    \item \textbf{0}: Not inclusive (The question contains biased or discriminatory language or assumptions). \textit{Example: ``Why are the women in the video acting emotionally?''}
    \item \textbf{1}: Slightly inclusive (The question is mostly neutral but could be phrased more inclusively). \textit{Example: ``What are the people in the video doing?'' (If the context strongly implies a specific gender)}
    \item \textbf{2}: Moderately inclusive (The question attempts to use neutral language but might still have some underlying assumptions). \textit{Example: ``What is the role of each person in the scene?''}
    \item \textbf{3}: Highly inclusive (The question uses neutral and respectful language, avoiding any biased or discriminatory assumptions about gender, race, age, etc.). \textit{Example: ``What actions are the individuals performing in the video?''}
\end{itemize}

\section*{Examples from Our Multihop Video Question Generation Dataset}

To illustrate the characteristics of the multihop questions within our newly created dataset MVQ-60, we present a selection of examples in Table~\ref{tab:multihop_examples}. Each row includes the constituent questions, their respective answers, the involved video clip names, and the resulting initial multihop question. It is important to note that these are the initial multihop (merged) questions generated by our system, after paraphrasing they often exhibit improved fluency and naturalness.

\textbf{Note:} Due to space constraints, we have presented a subset of the generated multihop questions. The full dataset contains a diverse range of questions requiring various forms of temporal and relational reasoning across different video segments from the TV show ``Friends''. The structure of these examples demonstrates how our dataset links information across potentially non-contiguous video clips through the composition of simpler questions

\begin{table*}[ht!]
    \centering
    \caption{Examples from Our Multihop Video Question Generation Dataset}
    \label{tab:multihop_examples}
    \scriptsize
    \begin{tabular}{p{2cm}p{1cm}p{2.25cm}p{2cm}p{2cm}p{2.25cm}p{3cm}}
        \toprule
        \textbf{Question 1} & \textbf{Answer 1} & \textbf{Video Clip 1} & \textbf{Question 2} & \textbf{Answer 2} & \textbf{Video Clip 2} & \textbf{Merged Question} \\
        \midrule
        Who was talking on the phone before Joey picked up the phone the first time? & Ross & friends, s02e01, seg02, clip, 07 & Who was Joey talking with when Ross went inside? & Joey was talking with his dad & friends, s02e01, seg02, clip, 21 & Who was Joey talking with when , the person Who was talking on the phone before Joey picked up the phone the first time?, went inside? \\
        \addlinespace
        Who was talking on the phone before Joey picked up the phone the first time? & Ross & friends, s02e01, seg02, clip, 07 & Where did Ross went after the conversation with Rachel? & Ross went inside the house & friends, s02e01, seg02, clip, 21 & Where did , the person Who was talking on the phone before Joey picked up the phone the first time?, went after the conversation with Rachel? \\
        \addlinespace
        Who does Charlie disagree knows art when Ross mentions him/her? & Joey & friends, s09e21, seg02, clip, 18 & Why does Joey joke with Ross after he gives suggestions for his date? & Joey jokes becasue Ross has detailed ideas specific to Joey's date's preferences. & friends, s09e21, seg02, clip, 08 & Why does , the person Who does Charlie disagree knows art when Ross mentions him/her?, joke with Ross after he gives suggestions for his date? \\
        \addlinespace
        Who does Charlie disagree knows art when Ross mentions him/her? & Joey & friends, s09e21, seg02, clip, 18 & Why doesn't Joey know what he just said after getting asked by Ross? & His brain is thinking about monster trucks & friends, s09e21, seg02, clip, 12 & Why doesn't , the person Who does Charlie disagree knows art when Ross mentions him/her?, know what he just said after getting asked by Ross? \\
        \addlinespace
        Who came to the room when Castle was talking? & Ryan & castle, s06e21, seg02, clip, 16 & Who comes looking for Ryan after he hangs up the phone? & Esposito comes looking for Ryan. & castle, s06e21, seg02, clip, 11 & Who comes looking for , the person Who came to the room when Castle was talking?, after he hangs up the phone? \\
        \addlinespace
        Who came to the room when Castle was talking? & Ryan & castle, s06e21, seg02, clip, 16 & What is Lanie waving around in her hand when she is facing Ryan and Esposito? & A pen. & castle, s06e21, seg02, clip, 18 & What is Lanie waving around in her hand when she is facing , the person Who came to the room when Castle was talking?, and Esposito? \\
        \addlinespace
        Who follows beckett out of montgomerys office after she leaves montgomerys office? & Castle & castle, s03e22, seg02, clip, 03 & What type of cup does Castle sit by when he clasps his hands? & Wine glass. & castle, s03e22, seg02, clip, 15 & What type of cup does , the person Who follows beckett out of montgomerys office after she leaves montgomerys office?, sit by when he clasps his hands? \\
        \addlinespace
        Who follows beckett out of montgomerys office after she leaves montgomerys office? & Castle & castle, s03e22, seg02, clip, 03 & Who started jumping onto Beckett after Castle opened the door? & Seeger & castle, s03e22, seg02, clip, 22 & Who started jumping onto Beckett after , the person Who follows beckett out of montgomerys office after she leaves montgomerys office?, opened the door? \\
        \bottomrule
    \end{tabular}
\end{table*}

 \begin{table*}[ht]
\centering
\caption{Qualitative examples of MVQG generations.}
\label{tab:mvqg_errors}
\scriptsize
\begin{tabular}{|p{2cm} | p{3.5cm} | p{3.5cm} | p{3.5cm} | p{3.5cm}|}

\toprule
\textbf{Series} & \textbf{Correct Generation} & \textbf{Multi-Hop Failure} & \textbf{External Knowledge Leakage} & \textbf{Hallucination} \\
\hline

BBT & What did , the person who wishes Sheldon a happy Valentines Day after he opens the door?, do after Sheldon told her he was being selfish? & What does the person who is eating with Sheldon say and who knocks on the door when Sheldon is talking? & What does , the person who is talking to Amy about moving in together after returning from Princeton?, say when they discuss their living arrangements? & What does , the person who gives Sheldon a Star Wars gift during his birthday party?, say when he opens the present? \\ \hline

Friends & What is , the person who came into the apartment when Leonard was on the phone?, holding when she is talking to Leonard? & What is the person who is sitting next to Monica doing and what does Rachel say when she enters the room? & What does , the person who is talking to Rachel when she says she got off the plane?, do after hearing her decision? & What does , the person who is cooking Thanksgiving dinner with Monica?, do when Joey accidentally drops the turkey? \\ \hline

HIMYM & What does , the person who is in charge when Howard and Sheldon work together?, do when he is talking to Howard and Raj? & What does the person who gives Ted a drink say and who walks into the apartment when Barney is talking? & What does , the person who is sitting in front of Ted when he starts telling the story about how he met their mother?, say when he gives them something? & What does , the person who plays the guitar during Robin’s farewell party?, say when Ted offers a toast? \\ \hline

Grey's Anatomy & What did , the person who is playing the piano when everyone is singing to Bernadette?, say when he was talking to Raj? & What does the patient who is lying on the bed do and who enters the room when Meredith is looking at the monitor? & What does , the person who is talking to Meredith before Derek's accident?, say when she expresses her concern? & What does , the person who argues with Cristina about the heart surgery?, do when the patient flatlines? \\ \hline

Castle & What did , the person who is wearing a brown trench coat when Beckett enters the alley?, say when she finds the second clue? & What is the person who found the evidence doing and what does Beckett say when she enters the office? & What does , the person who is confronting Castle when he finds out who killed his mother?, do when Castle reacts? & What does , the person who brings evidence to Beckett during the rooftop chase?, say when she finds the clue? \\ \hline

House & What does , the person who walks into the patient’s room after House finishes speaking with the nurse?, do when he notices the charts are missing? & What does the nurse who checks the IV bag say and what is House doing when he talks to Wilson? & What does , the person who is inside the house when House crashes his car into it?, say when House approaches? & What does , the person who challenges House’s diagnosis in the operating room?, do when the patient wakes up? \\ \hline

\end{tabular}
\end{table*}

\begin{table*}[h]
\centering
\caption{Examples of structured prompts and their outputs. These structured prompts often confused the model or led to verbose, incoherent generation.}
\label{tab:prompt_failures}
\begin{tabular}{|p{4.5cm}|p{8.5cm}|}
\hline
\textbf{Prompt} & \textbf{Generated Output} \\
\hline
\textit{“Think step by step. First, look at the first video. Then, look at the second. Now generate a question that connects both.”} & “Okay. First video: Sheldon is holding something. Second video: Raj is cooking. What is happening in both?” \newline \textit{(Fails to form a coherent multi-hop link)} \\
\hline
\textit{“Understand what happens in the first clip, then the second, and finally write a question that links them.”} & “In the first clip, someone is talking. In the second, there is a kitchen. Can you describe both?” \newline \textit{(Vague; no reasoning or question structure)} \\
\hline
\textit{“Let’s think carefully. Break down the events and create a multi-hop question based on them.”} & “Hmm. First event: someone holding sword. Second event: Raj in kitchen. What’s the question here?” \newline \textit{(Fails to produce a question; stays in meta-mode)} \\
\hline
\end{tabular}
\end{table*}

\subsection{Ablation Study}\label{app: Ablation Study}
\begin{table*}[t!]
    \centering 
    \scriptsize
    \caption{ Performance Comparison of Ablations}
    \begin{tabular}{lcccccc}
        \hline
        \textbf{Model} & \textbf{Fluency} & \textbf{Video Relevance} & \textbf{Multi-Hop Reasoning} & \textbf{Engagingness}&  \textbf{Factual Correctness} & \textbf{Inclusiveness}\\
        \hline
        Full Model & 2.81 & 2.91 & 2.85 & 2.75 & 2.92 & 2.78\\
        Without Video & 2.66 & 2.09 & 1.54 & 2.25 & 2.36 & 2.05 \\
        Single Component & 2.31 & 2.39 & 1.24 & 2.32 & 1.98 & 2.24\\
        \hline
    \end{tabular}
    \label{tab:ablation}
\end{table*}

To better understand the contributions of various components in \textbf{VideoChain}, we conducted an ablation study by modifying the model in two key ways: (1) removing the video embeddings to evaluate the reliance on textual information, and (2) simplifying the architecture into a single-component system to directly generate multi-hop questions. These experiments highlight the significance of both the multimodal input and the modular design in enhancing the model's performance on multi-hop question generation tasks. The results are shown in Table \ref{tab:ablation}

\noindent\textbf{Text-Only Model:}
In this configuration, we removed video embeddings and trained the model using only textual data—transcripts and prompts. This setup was designed to assess how much the model relies on visual information versus language alone when generating coherent and contextually grounded multi-hop questions.
Compared to the full model, the text-only version showed a drop across all metrics. Most notably, video relevance fell from 2.91 → 2.09, and multi-hop reasoning dropped from 2.85 → 1.54. While fluency remained relatively high (2.81 → 2.66), the absence of visual grounding led to decreased factual correctness (2.97 → 2.36) and overall inclusiveness (2.78 → 2.05). This suggests that visual information plays a critical role in enabling richer, more contextually grounded question generation.

\noindent\textbf{Single-Component Multi-Hop Generation (Direct Multihop Generation):}
In this variant, we removed the modular design and trained the model to generate multi-hop questions in a single stage, directly from paired video segments and transcripts. While this setting retained multimodal inputs, it lacked the iterative structure of the full model. As shown in Table~\ref{tab:ablation}, this led to a sharp drop in performance, particularly in \textit{multi-hop reasoning} (1.24 vs. 2.85) and \textit{factual correctness} (1.98 vs. 2.97). The model often produced concatenated or shallow questions, failing to perform genuine multi-step reasoning. This highlights the importance of structured decomposition for handling complex, compositional question generation.

\noindent\textbf{Evaluation and Comparative Metrics:}
Both ablation settings were evaluated using the same metrics as the full model: \textit{fluency}, \textit{relevance}, \textit{multi-hop reasoning complexity}, and \textit{factual correctness}. Let $S_{\text{full}}$, $S_{\text{text-only}}$, and $S_{\text{direct-mh}}$ denote the average scores for the full model, the text-only model, and the single-component model, respectively. The relative performance drop $\Delta S$ for each ablation is computed as:
\begin{equation}\scriptsize
    \Delta S_{\text{text-only}} = S_{\text{full}} - S_{\text{text-only}}, \quad \Delta S_{\text{direct-mh}} = S_{\text{full}} - S_{\text{direct-mh}}.
\end{equation}

We observe $\delta S_{\text{direct-mh}}$ to be 0.76 and $\delta S_{\text{text-only}}$ to be 0.69, the significant performance degradation in the \textbf{text-only} model, is due to tasks requiring visual grounding. Similarly, the \textbf{single-component model} is observed to underperform on tasks requiring complex multi-hop reasoning due to the absence of iterative refinement.
The results of ablation study shown in table~\ref{tab:ablation}, underscores the critical role of \textbf{video embeddings} in grounding questions in visual context and highlights the effectiveness of our model's \textbf{modular design} in handling multi-hop reasoning. 
\section{Qualitative Examples of MVQG Error Types}\label{app: error_analysis}

We conducted a detailed error analysis to understand the common failure modes in our MVQG framework. Table~\ref{tab:error} summarizes the frequency of each error category before and after applying mitigation strategies. Below, we elaborate on the nature of each error, provide examples, and describe how each was addressed.

\textbf{Multi-Hop Reasoning Failures}: The model often concatenated two zero-hop questions using ``and'' instead of forming a meaningful reasoning chain. \\
\textit{Incorrect}: ``What was Amy holding when she is talking to Penny \textbf{and} who was with Sheldon in the car?'' \\
\textit{Expected}: ``What was Amy holding when she is talking to the girl who was in the car with Sheldon?'' \\
We mitigated this by refining training prompts with clearer examples and post-processing rules, reducing the error rate from 24\% to 6\%.

\textbf{Factual Inaccuracy}: Some questions introduced incorrect or unsupported claims not grounded in the video.\textit{Example}: ``What did Rachel say when she got the job offer in Paris?'' (scene not present). \\
We applied negative sampling and semantic filters to reduce such cases from 12\% to 7\%.

\textbf{Semantic Drift}: The model occasionally produced abstract or off-topic questions.
\textit{Example}: ``How do characters reflect on their past experiences?'' \\
Prompt refinement and focused sampling during fine-tuning reduced this issue from 11\% to 5\%.

\textbf{Grammatical Errors}: Syntax or fluency issues such as tense mismatch or fragmented clauses.
\textit{Example}: ``What do Monica was cooking when Ross came in?'' \\
Due to consistent training and BART’s language generation strength, these errors decreased from 5\% to 3\%.

\textbf{Redundancy}: Some questions repeated the same concepts across hops.
\textit{Example}: ``Who was with Rachel and who was talking to Rachel in the kitchen?'' \\
We filtered such patterns post-merging, reducing redundancy from 9\% to 4\%.

\textbf{Ambiguity or Vagueness}: Questions lacked clarity or contained poorly grounded references. 
\textit{Example}: ``What did he do after she left?'' \\
This was addressed through coreference-aware sampling, reducing such errors from 14\% to 6\%.

\textbf{External Knowledge Leakage}: The model occasionally used memorized facts beyond input scope. 
\textit{Examples}: \\
``What does Sheldon say to Leonard before they move to their new apartment?'', 
``How does Ross react when Rachel leaves for Paris?'' \\
By constraining context and refining prompts, hallucination was reduced from 20\% to 8\%, and leakage from 32\% to 8\%.

Correct generations demonstrate well-formed multi-hop questions that refer to grounded visual-textual content and follow the long-form template described in Section~6.4. Multi-hop reasoning failures typically involve the incorrect use of conjunctions (e.g., “and”) instead of forming compositional reasoning chains. External knowledge leakage reveals the model's tendency to rely on memorized facts from the broader TV series rather than the provided scene. Hallucination refers to entirely fabricated content not grounded in the input video or known series lore.

These examples highlight the diversity of errors our model encounters and the necessity for strong grounding and compositional reasoning mechanisms in MVQG systems.
\subsection{Prompt Variants and Failure Cases}
As part of our baseline prompt engineering study, we investigated the impact of structured prompting strategies inspired by chain-of-thought reasoning. While such approaches have proven effective in text-based reasoning tasks, we found that they were not beneficial in our setting due to the combination of a smaller base model (BART-large) and the inherent complexity of the MVQG task.

Despite our initial expectations, prompts designed to simulate step-by-step reasoning often resulted in incoherent, vague, or incomplete outputs. Table~\ref{tab:prompt_failures} shows representative examples highlighting these failure cases.

These findings informed our final choice of using clear and direct prompts, such as: \textit{“Based on the two video segments and their transcripts, generate a question that requires integrating information from both videos.”} This prompt yielded more stable and contextually grounded outputs, particularly when paired with a well-trained zero-hop module. We include these prompt examples and results in the supplementary material for transparency and reproducibility.

\begin{table}[t!]
  \centering 
  \scriptsize
  \caption{ Improvements in Error categories}
  \begin{tabular}{lcc}
    \hline
    \textbf{Error Category} & \textbf{Error before} & \textbf{Error After} \\
    & \textbf{improvement (\%)} & \textbf{improvement (\%)} \\
    \hline
    Multihop Reasoning failure & 24 & 6 \\
    Factual Inaccuracy & 12 &  7\\
    Semantic Drift	& 11 & 5\\
    Grammatical Errors	& 5 & 3\\
    Redundancy & 9 & 4\\
    Ambiguity or Vagueness & 14 & 6\\	
    External Knowledge Leakage & 32 & 8\\
    \hline
  \end{tabular}
  \label{tab:error}
\end{table}

\begin{table*}[ht!]
    \centering
    \caption{Summary of Explored Video Question Answering Datasets}
    \label{tab:dataset_summary}
    \scriptsize
    \begin{tabular}{p{0.5cm}p{2cm}p{3cm}p{9cm}}
        \hline
        \textbf{No.} & \textbf{Dataset Name} & \textbf{Primary Focus} & \textbf{Short Description} \\
        \hline
        1 & MSR-VTT~\cite{xu2016msrvtt} & Open Domain Captioning & A large-scale dataset primarily for video captioning, containing 10,000 videos with 20 human-annotated captions per video. It's also used as a base for VQA tasks. \\
        2 & HowTo100M~\cite{miech2019howto100m} & Instructional Videos & A massive dataset of over 1 million narrated instructional videos, focusing on explaining how to perform various tasks. \\
        3 & TVQA~\cite{lei2018} & TV Shows (6) & A large-scale VideoQA dataset built upon 6 popular English-language TV shows, featuring multiple-choice questions and answers, along with subtitles and video frames. \\
        4 & ActivityNet-QA~\cite{yu2019activitynetqa} & Activity Understanding & Contains human-annotated question-answer pairs on videos from the ActivityNet dataset, designed to test models' long-term spatio-temporal reasoning abilities. \\
        5 & NExT-QA~\cite{xiao2021} & Explanation of Video & A VideoQA benchmark specifically created to evaluate the explanation of video content, requiring models to reason about causal and temporal relationships between actions and objects. \\
        6 & TGIF-QA~\cite{jang2017tgif} & Animated GIFs & Features question-answer pairs for animated GIFs from the TGIF dataset, suitable for evaluating video-based Visual Question Answering techniques on short, dynamic visual content. \\
        7 & MovieQA~\cite{tapaswi2016movieqa} & Movies & A dataset for question answering about movies, evaluating story comprehension from both video and textual sources like plot synopses and subtitles, with multiple-choice answers. \\
        8 & MVBench~\cite{li2024mvbench} & Temporal Understanding & A comprehensive benchmark designed to evaluate the temporal understanding capabilities of multimodal large language models (MLLMs) across 20 diverse dynamic video tasks. \\
        9 & MSRVTT-QA~\cite{xu2017video} & VQA on MSR-VTT & A benchmark for Visual Question Answering created based on the MSR-VTT video captioning dataset, evaluating the ability to answer questions grounded in video content described by captions. \\
        10 & MSVD-QA~\cite{xu2017video} & VQA from MSVD & A VideoQA dataset generated from the descriptive sentences in the MSVD dataset, providing a large set of question-answer pairs based on short video snippets and their textual descriptions. \\
        11 & TVQA+~\cite{lei2021tvqa} & Visual Grounding in TVQA & An extension of the TVQA dataset that includes detailed bounding box annotations, explicitly linking depicted objects to visual concepts mentioned in the questions and answers for enhanced visual grounding. \\
        12 & TGIF (Tumblr GIF)~\cite{li2016tgif} & GIF Descriptions & A dataset of 100,000 animated GIFs collected from Tumblr, each accompanied by several descriptive sentences provided by humans. \\
        13 & VideoInstruct~\cite{Maaz2023VideoChatGPT} & Video Instruction Following & A dataset comprising high-quality video and instruction pairs, used to train models like Video-ChatGPT to follow instructions presented in video format. \\
        14 & AGQA~\cite{grunde2021agqa} & Spatio-Temporal Reasoning & A benchmark designed to evaluate compositional spatio-temporal reasoning in videos, focusing on understanding actions, their attributes, and their relationships within a scene. \\
        15 & VALUE~\cite{li2021value} & General V\&L Understanding & A benchmark created to test the generalizability of video-and-language understanding models across a wide range of tasks, domains, and existing datasets. \\
        16 & How2QA~\cite{li2020hero} & QA on HowTo100M & A VideoQA dataset collected on the same videos as the HowTo100M dataset, featuring multiple-choice questions and answers related to the instructional content of the videos. \\
        17 & iVQA~\cite{liu2018ivqa} & Instructional Video QA & An open-ended VideoQA benchmark specifically for instructional videos, featuring multiple correct answer annotations for each question and requiring detailed video understanding. \\
        18 & IntentQA~\cite{li2023intentqa} & Social Activity Intents & A dataset focusing on the diverse intents behind actions observed in daily social activities, designed to evaluate models' ability to understand the underlying motivations in videos. \\
        19 & SUTD-TrafficQA~\cite{xu2021sutd} & Traffic Video QA & A dataset specifically focused on question answering related to traffic scenarios, requiring understanding of various events, objects, and their interactions within traffic videos. \\
        20 & STAR Benchmark~\cite{wu2024star} & Situated Reasoning & A benchmark aimed at evaluating how well models can capture and utilize present knowledge directly from the surrounding visual situations depicted in videos. \\
        21 & MSRVTT-MC~\cite{yu2018joint} & Multiple Choice VQA on MSR-VTT & A multiple-choice video question-answering dataset created based on the MSR-VTT dataset, offering a different evaluation format compared to open-ended QA. \\
        22 & TVBench~\cite{cores2024tvbench} & Temporal Understanding in VQA & A benchmark specifically created to evaluate the temporal understanding capabilities of VideoQA models, focusing on questions that require reasoning over time within the video. \\
        23 & Neptune~\cite{nagrani2024neptune} & Long Video QA & A dataset consisting of challenging question-answer-decoy sets for long-form videos (up to 15 minutes), pushing the limits of models' ability to understand and reason over extended video durations. \\
        24 & DramaQA~\cite{choi2021dramaqa} & Dialogue and Narrative Understanding & A dataset focused on two key aspects of video understanding: comprehending character dialogue and understanding the broader narrative flow in movies and TV series. \\
        25 & EgoTaskQA~\cite{jia2022egotaskqa} & Egocentric Video QA & A benchmark containing balanced question-answer pairs for egocentric (first-person perspective) videos, primarily focusing on understanding the actions performed by the camera wearer. \\
        26 & Perception Test~\cite{patraucean2023perception} & Perception and Reasoning & A benchmark designed to evaluate the fundamental perception and reasoning skills of multimodal models when processing video content. \\
        27 & VLEP~\cite{lei2020more} & Video-and-Language Event Prediction & Contains examples of future event prediction in videos along with their textual rationales, testing the ability to anticipate what will happen next based on the observed context. \\
        28 & KnowIT VQA~\cite{garcia2020knowit} & QA on The Big Bang Theory & A video dataset with human-generated question-answer pairs specifically centered around the content and characters of the popular TV show ``The Big Bang Theory,'' focusing on domain-specific knowledge. \\
        29 & MovieFIB~\cite{maharaj2017dataset} & Movie Fill-in-the-Blank & A benchmark featuring fill-in-the-blank style questions based on detailed descriptive video annotations created for the visually impaired, testing fine-grained visual understanding. \\
        30 & HowToVQA69M~\cite{yang2021just} & Large-Scale HowTo VQA & A very large-scale VideoQA dataset built upon the HowTo100M video dataset, containing approximately 69 million question-answer pairs. \\
        31 & TutorialVQA~\cite{colas2019tutorialvqa} & Answer Spans in Tutorials & A dataset designed for the task of finding specific answer spans within the transcripts of tutorial videos, with questions and manually collected answer spans. \\
        32 & Video Localized Narratives~\cite{Voigtlaender23CVPR} & Vision and Language Connection & A dataset that explicitly connects vision and language by providing detailed, localized narratives describing objects and actions within specific regions of video frames. \\
        33 & CRIPP-VQA~\cite{patel2022cripp} & Counterfactual Reasoning & A dataset designed to evaluate counterfactual reasoning about implicit physical properties through video question answering, requiring models to understand ``what if'' scenarios. \\
        34 & CausalChaos!~\cite{parmar2024causalchaos} & Causal Video QA & A dataset specifically created for evaluating causal reasoning in video question answering, using animated content from Tom and Jerry cartoons to focus on cause-and-effect relationships. \\
        35 & CinePile~\cite{rawal2024cinepile} & Long Video QA (Movies) & A question-answering-based dataset focused on the understanding of long-form video content, specifically utilizing movie data. \\
        36 & RoadTextVQA~\cite{tom2023reading} & Text Understanding in Driving Videos & A dataset focused on the task of understanding text and signs present in videos captured from driving scenarios, crucial for applications in autonomous driving. \\
        37 & Social-IQ 2.0~\cite{zadeh2019social} & Social Intelligence in Video & A dataset designed to evaluate the social intelligence of AI models in understanding videos, focusing on social interactions, nonverbal cues, and human behavior. \\
        38 & VCG+112K~\cite{maaz2024videogpt+} & Video Instruction Following & Another large-scale dataset for video instruction following, containing over 112,000 video-instruction pairs for training models to execute tasks based on video instructions. \\
        39 & WildQA~\cite{castro2022wildqa} & VQA in Outside Settings & A video understanding dataset comprising videos recorded in unconstrained, real-world outside environments, designed to evaluate the robustness of models in more natural settings. \\
        40 & Vript~\cite{yang2024vript} & Fine-Grained Video-Text & A fine-grained video-text dataset featuring high-resolution videos and detailed, rich annotations, aiming for a deeper understanding of the relationship between visual and textual information in videos. \\
        \hline
    \end{tabular}
\end{table*}

\end{document}